\documentclass{article}

\usepackage[nonatbib,final]{idiap_2020}

\usepackage[utf8]{inputenc} % allow utf-8 input
\usepackage[T1]{fontenc}    % use 8-bit T1 fonts
\usepackage{hyperref}       % hyperlinks
\usepackage{url}            % simple URL typesetting
\usepackage{booktabs}       % professional-quality tables
\usepackage{amsfonts}       % blackboard math symbols
\usepackage{nicefrac}       % compact symbols for 1/2, etc.
\usepackage{microtype}      % microtypography
\usepackage{booktabs,multirow, multicol}
\usepackage{fullpage}
\usepackage{enumitem}
\usepackage{graphicx}
\usepackage{longtable}
\usepackage{makecell}
\usepackage{titlesec}
\usepackage{array}
\usepackage{authblk}

\usepackage[caption=false]{subfig}

\usepackage{enumitem}

\usepackage{rotating}

\usepackage{boldline}
\title{The High-Quality Wide Multi-Channel Attack (HQ-WMCA) database}

  \author{Zohreh Mostaani,
        Anjith George,
        Guillaume~Heusch,
        David Geissb\"uhler,
        and~S\'ebastien~Marcel,~\textit{Senior~Member,~IEEE}% <-this % stops a space
\thanks{ Zohreh Mostaani, Anjith George, Guillaume Heusch, David Geissb\"uhler, and~S\'ebastien Marcel are with the
  Idiap Research Institute, Switzerland, e-mails: \{zohreh.mostaani, anjith.george, guillaume.heusch, david.geissbuhler, sebastien.marcel\}@idiap.ch}% <-this % stops a space
}

\begin{document}

\maketitle

\begin{abstract}
The High-Quality Wide Multi-Channel Attack database (HQ-WMCA) database extends the previous Wide Multi-Channel Attack database(WMCA) \cite{george_mccnn_tifs2019}, with more channels
including color, depth, thermal, infrared (spectra), and short-wave infrared (spectra), and also a wide variety of attacks.

\end{abstract}

\section{Introduction}

Though face recognition systems are achieving near-perfect accuracies, face recognition systems remain vulnerable to presentation attacks. Several different methods have been proposed to overcome this vulnerability \cite{george-icb-2019,handbook2}. Metrics for evaluating the methods are also standardized \cite{ISO,ISO1}.
Recent research has highlighted the importance of multi-channel information for robust face presentation attack detection \cite{george_mccnn_tifs2019,george-tifs-2020,nikisins2019domain,george2020face}. However, there is a wide range of channels available to PAD. To advance the research in multi-channel PAD, we make available the HQ-WMCA large multi-channel dataset covering a wide range of 2D, 3D, and partial attacks.

\section{High-Quality Wide Multi-Channel Attack (HQ-WMCA) database}

The High-Quality Wide Multi-Channel Attack (HQ-WMCA) database \cite{heusch2020deep} consists of 2904 short multi-modal video recordings of both bona-fide and presentation attacks. There are 555 bonafide presentations from 51 participants and the remaining 2349 are presentation attacks. The data is recorded from several channels including color, depth, thermal, infrared (spectra), and short-wave infrared (spectra).

Preprocessed images for some of the channels are also provided for part of the data used in the publication.

The HQ-WMCA database is produced at Idiap within the framework of "IARPA BATL" project \cite{ODIN} and it is intended for research, development, and testing in biometrics and biomedical analysis.

\begin{figure}[ht]
    \centering
    \includegraphics[width=0.7\linewidth]{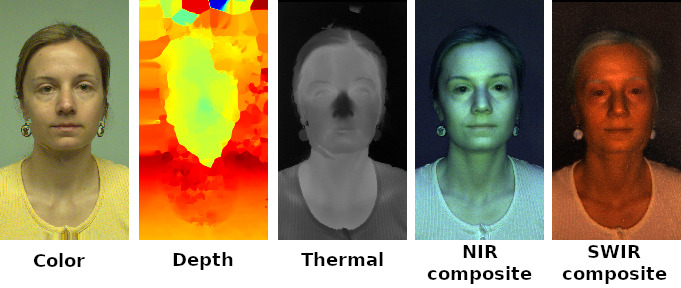}
    \caption{Sample images of bonafide from the database for different channels. From left to right the images are from color, depth, thermal, NIR, and SWIR channels. NIR composite and SWIR composite are obtained by combining a subset of the images from the corresponding spectra.}
    \label{fig:bf_sample}
\end{figure}

\newpage
If you use this database, please cite the following publication \cite{heusch2020deep}.

\section{Database Description:}

The HQ-WMCA  includes short video recordings from different channels including color, depth, thermal, near infra-red (NIR) and short wave infra-red (SWIR).

The sensors used for the data acquisition are:
\begin{itemize}
    \item Basler acA1920-150uc: This sensor records the data for color channel.
    \item Basler acA1920-150um: This sensor records the data for NIR channel. Two identical sensors were mounted on the left and right of the system to capture data for stereo reconstruction purpose. These sensors are horizontally aligned and the distance between their centers is approximately 15 cm.
    \item Xenics Bobcat-640-GigE: This sensor records the data for SWIR channel.
    \item Xenics Gobi-640-GigE: This sensor records data from thermal channel.
    \item Intel Realsense D415: This sensor can records data from color, depth and infrared channels, however only the data from depth channel has been recorded.
\end{itemize}

Information about each sensor is described in Table. \ref{tab:sensor-info}

\begin{table}[h]
\centering
\caption{Sensor description for HQ-WMCA data}
\label{tab:sensor-info}
\begin{tabular}{cccc}
\specialrule{.1em}{.05em}{.05em}
Sensor name            & Channel & Resolution       & Frame rate (fps) \\
\specialrule{.1em}{.05em}{.05em}
Basler acA1920-150uc   & Color   & 1920$\times$1200 & 30               \\ \hline
Basler acA1920-150um   & NIR     & 1920$\times$1200 & 90               \\ \hline
Xenics Bobcat-640-GigE & SWIR    & 640$\times$512   & 90               \\ \hline
Xenics Gobi-640-GigE   & Thermal & 640$\times$480   & 30               \\ \hline
Intel Realsense D415   & Depth   & 720$\times$280   & 30               \\
\specialrule{.1em}{.05em}{.05em}
\end{tabular}
\end{table}

{}
Four LED modules are used for illumination besides the ambient illumination available in the room. Each LED module consists of LEDs operating in different wavelength mostly in NIR and SWIR. The used frequencies are 735, 850, 940, 1050, 1200, 1300, 1450, 1550, and 1650 nm. The NIR and SWIR camera captures data at the same time as a corresponding wavelength LED is illuminating. Therefore there are multiple streams of data corresponding to different illumination wavelength at the end of data acquisition for NIR and SWIR cameras. The data from cases when there is no LED illumination is captured. For stereo reconstruction from two NIR cameras separate data streams were recorded .An LED with wavelength of 735 nm is used for illumination in this case. It is showed in the cycle as stereo. The data from the other channels (color, depth, and thermal) is captured regardless of LED illumination and for whole duration of the data collection.

The data is collected during 2 seconds which is equal to 20 cycles of LED illumination. Each illumination cycle lasts for 100 ms. The trigger sequence for different cameras with respect to one LED cycle is showed in the Table. \ref{tab:led-info}.

\begin{sidewaystable}[h]
\centering
\caption{LED cycle description for synchronized sensors}
\label{tab:led-info}
\resizebox{\textwidth}{!}{
\begin{tabular}{llllllllllllllllllll}
\specialrule{.1em}{.05em}{.05em}
Sensor name   & Slot-01  & slot-02  & slot-03  & slot-04  & slot-05  & slot-06 &
 slot-07      & slot-08  & slot-09  & slot-10  & slot-11  & slot-12  & slot-13    & slot-14    & slot-15  & slot-16   & slot-17  & slot-18  & slot-19 \\
\specialrule{.1em}{.05em}{.05em}
&  & &  & &   &     &      &      &  & &  & &  & &  & &  & &    \\
\clineB{2-2}{4} \clineB{8-8}{4} \clineB{14-14}{4} \clineB{20-20}{4}
\multicolumn{1}{lV{4}}{Basler acA1920-150uc}   & \multicolumn{1}{lV{4}}{T} &  &   &  &  & \multicolumn{1}{cV{4}}{}      & \multicolumn{1}{lV{4}}{T} & &  & &  & \multicolumn{1}{lV{4}}{} & \multicolumn{1}{lV{4}}{T} & &  & &  & \multicolumn{1}{lV{4}}{} &        \\
 \clineB{3-7}{4} \clineB{9-13}{4} \clineB{15-19}{4}
 &   &  &   &  &   &   &    &       &   &  &   &  &   &  &   &  &   &   &        \\
 \clineB{2-2}{4} \clineB{4-4}{4} \clineB{6-6}{4} \clineB{8-8}{4} \clineB{10-10}{4} \clineB{12-12}{4} \clineB{14-14}{4} \clineB{16-16}{4} \clineB{18-18}{4} \clineB{20-20}{4}
\multicolumn{1}{lV{4}}{Basler acA1920-150um}   & \multicolumn{1}{lV{4}}{T} & \multicolumn{1}{lV{4}}{} & \multicolumn{1}{lV{4}}{T} & \multicolumn{1}{lV{4}}{} & \multicolumn{1}{lV{4}}{T}     & \multicolumn{1}{lV{4}}{}      & \multicolumn{1}{lV{4}}{T}     & \multicolumn{1}{lV{4}}{}      & \multicolumn{1}{lV{4}}{T} & \multicolumn{1}{lV{4}}{} & \multicolumn{1}{lV{4}}{T} & \multicolumn{1}{lV{4}}{} & \multicolumn{1}{lV{4}}{T} & \multicolumn{1}{lV{4}}{} & \multicolumn{1}{lV{4}}{T} & \multicolumn{1}{lV{4}}{} & \multicolumn{1}{lV{4}}{T} & \multicolumn{1}{lV{4}}{} &        \\ \clineB{3-3}{4} \clineB{5-5}{4} \clineB{7-7}{4} \clineB{9-9}{4} \clineB{11-11}{4} \clineB{13-13}{4} \clineB{15-15}{4} \clineB{17-17}{4} \clineB{19-19}{4}
 &   &  &   &  &       &       &       &       &   &  &   &  &   &  &   &  &   &  &        \\
  \clineB{2-2}{4} \clineB{4-4}{4} \clineB{6-6}{4} \clineB{8-8}{4} \clineB{10-10}{4} \clineB{12-12}{4} \clineB{14-14}{4} \clineB{16-16}{4} \clineB{18-18}{4} \clineB{20-20}{4}
\multicolumn{1}{lV{4}}{Xenics Bobcat-640-GigE} & \multicolumn{1}{lV{4}}{T} & \multicolumn{1}{lV{4}}{} & \multicolumn{1}{lV{4}}{T} & \multicolumn{1}{lV{4}}{} & \multicolumn{1}{lV{4}}{T}     & \multicolumn{1}{lV{4}}{}      & \multicolumn{1}{lV{4}}{T}     & \multicolumn{1}{lV{4}}{}      & \multicolumn{1}{lV{4}}{T} & \multicolumn{1}{lV{4}}{} & \multicolumn{1}{lV{4}}{T} & \multicolumn{1}{lV{4}}{} & \multicolumn{1}{lV{4}}{T} & \multicolumn{1}{lV{4}}{} & \multicolumn{1}{lV{4}}{T} & \multicolumn{1}{lV{4}}{} & \multicolumn{1}{lV{4}}{T} & \multicolumn{1}{lV{4}}{} &        \\ \clineB{3-3}{4} \clineB{5-5}{4} \clineB{7-7}{4} \clineB{9-9}{4} \clineB{11-11}{4} \clineB{13-13}{4} \clineB{15-15}{4} \clineB{17-17}{4} \clineB{19-19}{4}
&  &  &   &  &   &  &   &   &   &   &   &   &   &   &    &   &    &    &        \\
\clineB{2-2}{4} \clineB{8-8}{4} \clineB{14-14}{4} \clineB{20-20}{4}
\multicolumn{1}{lV{4}}{Xenics Gobi-640-GigE}   & \multicolumn{1}{lV{4}}{T} &   &    &   &  & \multicolumn{1}{lV{4}}{}      & \multicolumn{1}{lV{4}}{T}     &    &    &   &   & \multicolumn{1}{lV{4}}{} & \multicolumn{1}{lV{4}}{T} &   &   &   &  & \multicolumn{1}{lV{4}}{} &        \\
\clineB{3-7}{4} \clineB{9-13}{4} \clineB{15-19}{4}
 &   &  &   &  &   &   &   &   &   &  &   &  &   &  &   &  &   &   &        \\ \hline
LEDs SWIR & dark &     & 940  &     & 1050     &   & 1200 &   & 1300 &  & 1450 &     & 1550 &     & 1650 &   & dark &     &        \\ \hline
LEDs NIR   & ster&    & 940 &    & 1050    &  & stereo  &  & 735 &    & dark&    & stereo&    & 850 &    & dark&    &     \\ \hline
time {[}ms{]} & 0  &   & 11 &   & 22  &    & 33   &   & 44 &   & 55 &   & 66 &   & 77 &   & 88 &   & 100    \\
\specialrule{.1em}{.05em}{.05em}
\end{tabular}}
\end{sidewaystable}

In the table where there is "T" it means that the corresponding camera is triggered and the the data is captured in that time slot. The table only shows one cycle of illumination. This cycle repeats for 20 times and at the end there are different number of frames captured for each camera and each wavelength. It should be mentioned that all the sensors capture data synchronously except for "Intel Realsense D415". It was not possible for "Intel Realsense D415" to be triggered using the controller similar to other sensors. This sensor captures data for slightly more than the duration of the data capture to make sure that there is frames available when other sensors are triggered, and therefore the number of frames captured is more than others. The total number of frames acquired is described in the Table. \ref{tab:frame-info}.

\begin{table}[h]
\centering
\caption{Frame description for each data collection sample}
\label{tab:frame-info}
\begin{tabular}{cccc}
\specialrule{.1em}{.05em}{.05em}
Sensor                 & Channel & Wavelength (nm) & number of frames \\
\specialrule{.1em}{.05em}{.05em}
Basler acA1920-150uc   & Color   & NA              & 60               \\ \hline
Basler acA1920-150um   & NIR     & 735             & 20               \\ \hline
Basler acA1920-150um   & NIR     & 850             & 20               \\ \hline
Basler acA1920-150um   & NIR     & 940             & 20               \\ \hline
Basler acA1920-150um   & NIR     & 1050            & 20               \\ \hline
Basler acA1920-150um   & NIR     & dark            & 40               \\ \hline
Basler acA1920-150um   & NIR     & stereo          & 60               \\ \hline
Xenics Bobcat-640-GigE & SWIR    & 940             & 20               \\ \hline
Xenics Bobcat-640-GigE & SWIR    & 1050            & 20               \\ \hline
Xenics Bobcat-640-GigE & SWIR    & 1200            & 20               \\ \hline
Xenics Bobcat-640-GigE & SWIR    & 1300            & 20               \\ \hline
Xenics Bobcat-640-GigE & SWIR    & 1450            & 20               \\ \hline
Xenics Bobcat-640-GigE & SWIR    & 1550            & 20               \\ \hline
Xenics Bobcat-640-GigE & SWIR    & 1650            & 20               \\ \hline
Xenics Bobcat-640-GigE & SWIR    & dark            & 40               \\ \hline
Xenics Gobi-640-GigE   & Thermal & NA              & 60               \\ \hline
Intel Realsense D415   & Depth   & NA              & 70               \\
\specialrule{.1em}{.05em}{.05em}
\end{tabular}
\end{table}

All the sensors are mounted and fixed on a tripod according to the TREX design from TOR labs. The cameras are placed at a distance between 50-60cm from the subject face. The distance is shown on the data capture GUI.

Each subject is given an ID number upon arriving. This ID number is used during all the data capture, including bonafide and presentation attacks if necessary. This number is referred to as ``client\_id'' in case of bonafide and ``presenter\_id'' in case the subject is presenting an attack to the system.

In order to have more variability in the dataset the data acquisition is performed during three sessions. The sessions are different based on their illumination. The background in all the sessions is simple white.

\begin{itemize}
    \item \textbf{session 1:} Ambient office light illumination. There is no SWIR wavelength in the ambient illumination.
    \item \textbf{session 2:} Ambient office light illumination and a halogen lamp that is switched on far away from the cameras. This halogen lamp produces SWIR wavelength that illuminates the subject indirectly.
    \item \textbf{session 3:} There is no ambient office light illumination and two LED spot lights are illuminating from left and right. There is no SWIR wavelength in the spot lights.
\end{itemize}

The data includes bonafide and presentation attacks performed by the subjects participated in the data collection as well as the presentation attacks on the stand. Since the duration of the data collection was only 2 seconds, data was mostly captured twice from each presentations to include more samples. The file names for the two presentations are the same except for an 8 digit random sequence of numbers and characters at the end of the file. This part of the file name is called ``trial\_id''.
%-----------
\section{Presentations:}
\subsection{Bonafide:}

The bonafide data was captured from subjects while they were seated in front of the cameras with a neutral facial expression. The subjects were asked to remove their medical glasses if they wore one. The data from the subjects wearing the medical glasses was also captured.

\subsection{Presentation Attacks:}\label{pa groups}

The attacks for the BATL data acquisition is including ten categories of attacks and each of those categories may include sub-categories. This makes it easy to allocate the data properly to different subsets for training and testing models.

\begin{itemize}
    \item \textbf{Glasses (type 01)}: This type of attacks includes the following sub types.
    \begin{itemize}
        \item \textbf{00}: Retro glasses
        \item \textbf{01}: Funny eyes glasses
        \item \textbf{02}: Paper glasses
    \end{itemize}
    \item \textbf{Mannequin (type 02)}: This type of attacks includes the following sub types.
    \begin{itemize}
        \item \textbf{00}: Not specific
    \end{itemize}
    \item \textbf{Print (type 03)}: This type of attacks includes the following sub types.
    \begin{itemize}
        \item \textbf{00}: Printed photo on Matte paper using Typical office laser printer (CX c224e).
        \item \textbf{01}: Printed photo on Glossy paper using Typical office laser printer (CX c224e).
        \item \textbf{02}: Printed photo on Matte paper using Professional quality printer (Epson\_XP-860\_series instead of Canon 6010).
        \item \textbf{03}: Printed photo on Glossy paper using Professional quality printer (Epson\_XP-860\_series instead of Canon 6010).
    \end{itemize}
    \item \textbf{Replay (type 04)}: This type of attacks includes the following sub types.
    \begin{itemize}
        \item \textbf{00}: Video while played.
        \item \textbf{01}: Video while paused.
        \item \textbf{02}: Digital photo presented on an electronic display.
    \end{itemize}
    \item \textbf{Rigid mask (type 05)}: This type of attacks includes the following sub types.
    \begin{itemize}
        \item \textbf{00}: Non-transparent plastic mask.
        \item \textbf{01}: Transparent plastic mask without makeup.
        \item \textbf{02}: Transparent plastic mask with makeup.
        \item \textbf{03}: That'sMyFace resin mask.
        % \item \textbf{04}: That'sMyFace resin masks type 2 (This type is not used in the data collection. The data is there for completeness).
    \end{itemize}
    \item \textbf{Flexible mask (type 06)}: This type of attacks includes the following sub types.
    \begin{itemize}
        \item \textbf{00}: Full face silicon mask
        \item \textbf{01}: Half face silicon mask
    \end{itemize}
    \item \textbf{Paper mask (type 07)}: This type of attacks includes the following sub types.
    \begin{itemize}
        \item \textbf{00}: Not specific
    \end{itemize}
    \item \textbf{Wigs (type 08)}: This type of attacks includes the following sub types.
    \begin{itemize}
        \item \textbf{00}: Not specific
    \end{itemize}
    \item \textbf{Tattoo (type 09)}: This type of attacks includes the following sub types.
    \begin{itemize}
        \item \textbf{00}: Maori tribal face tattoo
    \end{itemize}
    \item \textbf{Makeup (type 10)}: This type of attacks includes the following sub types.
    \begin{itemize}
        \item \textbf{00}: Heavy contour makeup level 0.
        \item \textbf{01}: Heavy contour makeup level 1.
        \item \textbf{02}: Heavy contour makeup level 2.
        \item \textbf{03}: Pattern makeup level 0.
        \item \textbf{04}: Pattern makeup level 1.
        \item \textbf{05}: Pattern makeup level 2.
        \item \textbf{06}: Transformation makeup.
        \item \textbf{07}: Beauty makeup level 0.
        \item \textbf{08}: Beauty makeup level 1.
        \item \textbf{09}: Beauty makeup level 2.
    \end{itemize}
\end{itemize}

\begin{figure*}[ht]
\centering
  \subfloat[]{\includegraphics[height=2.5cm]{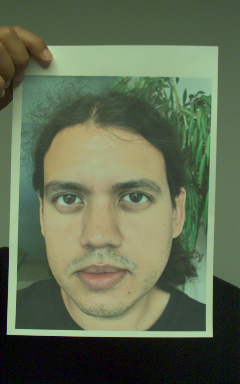}}%
\hfil
  \subfloat[]{\includegraphics[height=2.5cm]{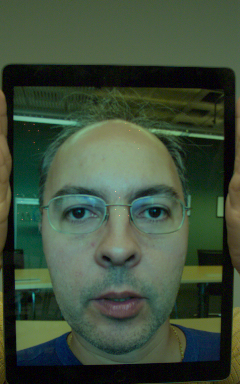}}%
\hfil
  \subfloat[]{\includegraphics[height=2.5cm]{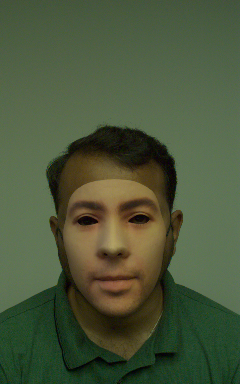}}%
\hfil
  \subfloat[]{\includegraphics[height=2.5cm]{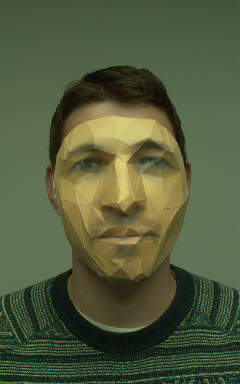}}%
\hfil
  \subfloat[]{\includegraphics[height=2.5cm]{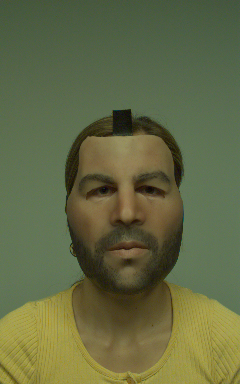}}%
\hfil
  \subfloat[]{\includegraphics[height=2.5cm]{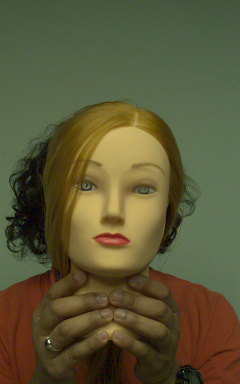}}%
\hfil
  \subfloat[]{\includegraphics[height=2.5cm]{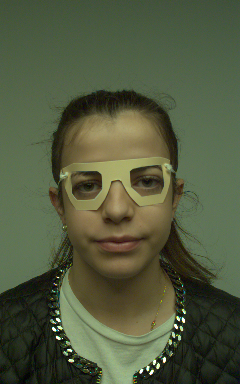}}%
\hfil
  \subfloat[]{\includegraphics[height=2.5cm]{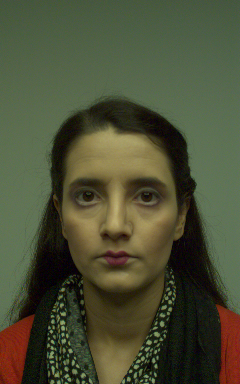}}%
\hfil
  \subfloat[]{\includegraphics[height=2.5cm]{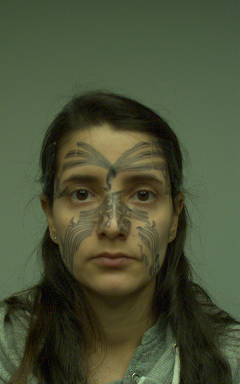}}%
\hfil
  \subfloat[]{\includegraphics[height=2.5cm]{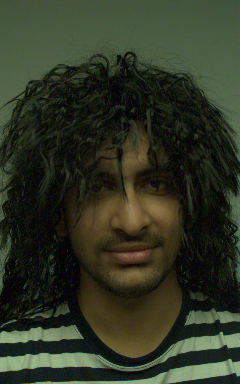}}%
  \caption{Example of attacks present in the database. (a) Print, (b) Replay, (c) Rigid mask, (d) Paper mask, (e) Flexible mask, (f) Mannequin,
          (g) Glasses, (h) Makeup, (i) Tattoo and (j) Wig.
          Note that only one particular example for each category is shown here, but there exists more variation across the database.
          For instance, print attacks have been crafted using different printers and different papers.}
\label{fig:attacks}
\end{figure*}

\begin{figure}[ht]
    \centering
    \includegraphics[width=0.6\linewidth]{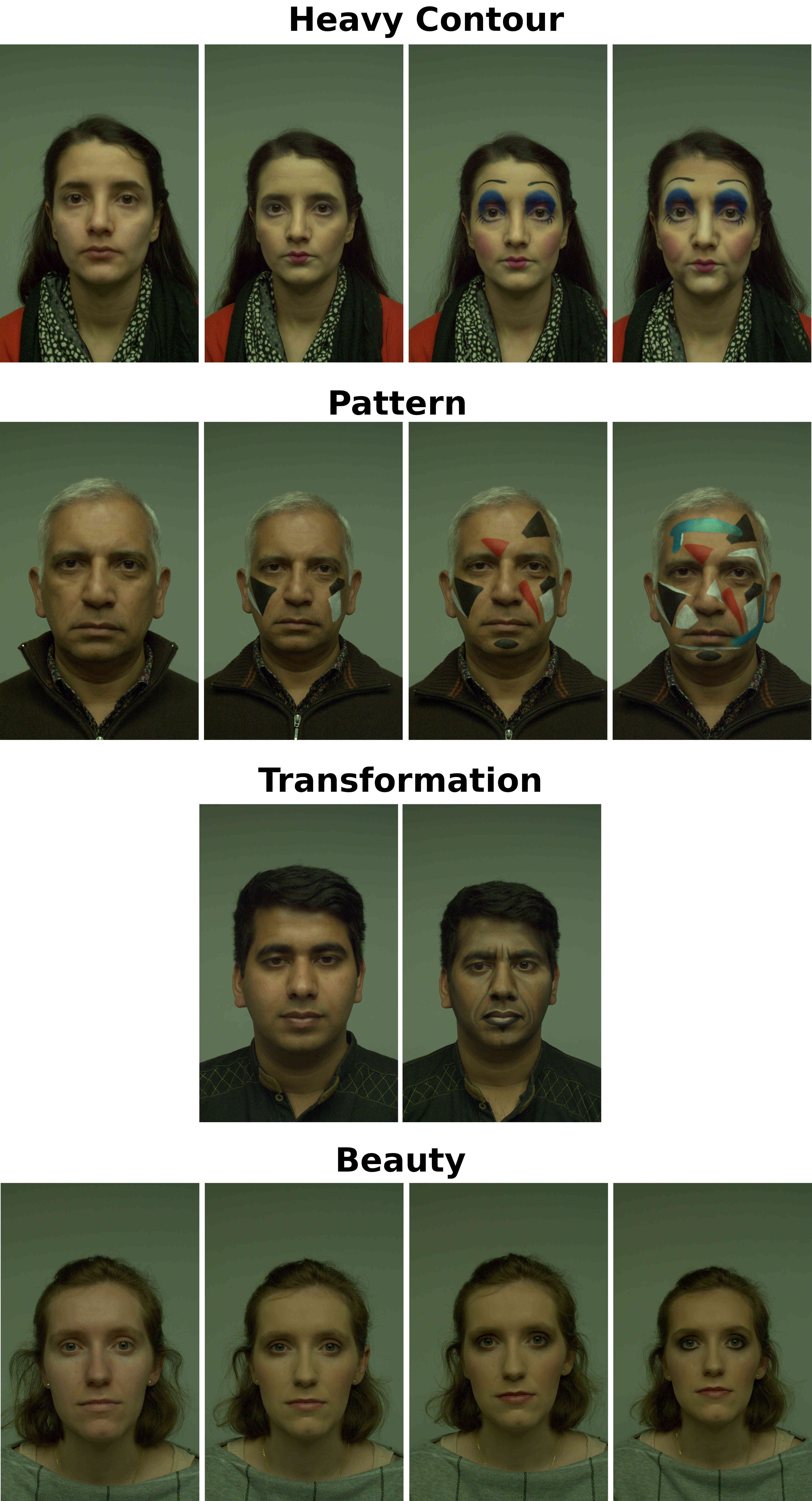}
    \caption{Sample images of makeup attacks for different sub-categories and levels. The images from top to bottom belongs to Heavy contour, Pattern, Transformation, and Beauty makeup. In each row the first image on the left is bonafide and from left to right the level of the makeup increases.}
    \label{fig:makeups}
\end{figure}

Examples of attacks from each category and for different sub-categories of makeup is shown in Fig. \ref{fig:attacks} and Fig. \ref{fig:makeups} respectively.

Each group has a ``type\_id'' and each sub group has a ``sub\_type\_id''. The presentation attack instruments (PAI) included in each sub group has a unique ``pai\_id'' as well as a ``client\_id'' which may not be unique. More detailed information about these numbers will be presented in the following.

\section{File Naming Convention:}
The file names in the database encode some information about the type of data they contain, for example bonafide or presentation attacks. Each saved file has the following name format: \\
\textbf{$<$site\_id$>$\_$<$session\_id$>$\_$<$client\_id$>$\_$<$presenter\_id$>$\_$<$type\_id$>$\_$<$sub\_type\_id$>$\_$<$pai\_id$>$}.\textbf{hdf5}

\begin{itemize}
    \item \textbf{site\_id:} The number represents the place of the data collection. In this database this number is always `1'.
    \item \textbf{session\_id:} The number associated to a session as mentioned before.
    \item \textbf{client\_id:} This number presents the identity of what is presented to the system. For bonafide, it is the ID given to the participant upon arrival and for the attacks, it is a number given to a PAI in this protocol. Please note that if the identity of a subject is the same as the identity of an attack this number is the same for both cases. One example is the silicon masks. If a silicon mask is made from subject `x' and subject `x' also participated as bonafide in the data collection the ``client\_id'' for bonafide and silicon mask is the same.
    \item \textbf{presenter\_id:} If a subject is presenting an attack to the system, this number is the subject's ``client\_id''. If the attack is presented on a support, this number is `0000'. If the capture is for bonafide this number is `0000' as well since there is no presenter in this case.
    \item \textbf{type\_id:} The attack types mentioned in \ref{pa groups}. For bonafide this number is `00'.
    \item \textbf{sub\_type\_id:} The sub\_types for each attack type mentioned in \ref{pa groups}. For bonafide  without glasses this number is `00' and if they wore medical glasses this number is '01'.
    \item \textbf{pai\_
    id:} The unique number associated with each and every PAI. This number for bonafide both with and without medical glasses is `000'.
\end{itemize}

Here are some examples for more clarification:

\begin{itemize}
    \item 1\_01\_0035\_0000\_00\_00\_000 : This is the \textbf{bonafide} file of client number 35 in session number 1 when they did not wear medical glasses.
    \item 1\_01\_0005\_0000\_00\_01\_000 : This is the \textbf{bonafide} file of client number 5 in session number 1 when they did wear medical glasses.
    \item 1\_02\_0109\_0001\_06\_01\_016 : This is a \textbf{flexible mask} attack where the identity 109 is presented to the camera by client number 1 in session 2.
    \item 1\_03\_0018\_0000\_03\_02\_002 : This is a \textbf{photo} attack with sub\_type two (Printed photo on Matte paper using Professional quality printer) where the identity 18 is presented to the camera using stand in session 03.
\end{itemize}

\section{Evaluation:}

The SWIR spectra is mainly used in the reference publication. Since the consecutive frames are correlated, only 10 frames from each video were selected. The frames are uniformly sampled in the temporal domain. The total number of 2904 presentations including bonafide and presentation attacks were grouped into three subsets, train, dev, and eval. The data split is done ensuring almost equal distribution of PA categories and disjoint set of client identifiers in each set. Each of the PAIs had different client id. The split is done in such a way that a specific PA instrument will appear in only one set. A low level database interface is implemented that handles loading and spatial and temporal alignment of the data \footnote{\url{https://gitlab.idiap.ch/bob/bob.db.hqwmca}}. The preprocessed method is described in details in the reference publication \cite{heusch2020deep} and the implementation is available publicly \footnote{\url{https://gitlab.idiap.ch/bob/bob.paper.pad_mccnns_swirdiff}}. The number of presentations in train, dev, and eval subset for the protocols in the reference publication is mentioned in Table \ref{tab:attacks-distribution}. Note that these are the numbers of the original videos fed to the pipeline, however the actual number of preprocessed images provided here are less than the numbers in the Table \ref{tab:attacks-distribution} due to failure of the face detection stage.

\begin{table}[h]
  \centering
  \caption{Number of examples for bonafide and attack examples in each set. The number of different identities
  is given in parenthesis. Note that having different identities provides variability for bonafide examples.}
  \begin{tabular}{llll}
    & \textbf{Train} & \textbf{Validation} & \textbf{Test}\\
  \midrule
    \textit{Bonafide} & 228 (21) & 145 (14) & 182 (16)\\
    Attacks & 742 & 823 & 784\\
  \bottomrule
  \end{tabular}
  \label{tab:set-stats}
\end{table}

\begin{table}
  \centering
  \caption{Distribution of attacks in the different sets, grandtest protocol is the superset of these two protocols}
  \begin{tabular}{llccc}
    & \textbf{Attack type} & \textbf{Train} & \textbf{Validation} & \textbf{Test}\\
  \midrule
    \multirow{7}{*}{\textbf{Impersonation}} & Print         & 48  & 98  & 0\\
                                            & Replay        & 36  & 100 & 126\\
                                            & Rigid Mask    & 162 & 118 & 140\\
                                            & Paper Mask    & 28  & 24  & 49\\
                                            & Flexible Mask & 90  & 86  & 48\\
                                            & Mannequin     & 20  & 38  & 77\\
                                            & \it{Total}    & \it{384} & \it{464} & \it{440}\\
  \midrule
    \multirow{5}{*}{\textbf{Obfuscation}} & Glasses     & 56  & 38  & 36\\
                                          & Makeup      & 264 & 271 & 258\\
                                          & Tattoo      & 24  & 24  & 24\\
                                          & Wig         & 14  & 26  & 26\\
                                          & \it{Total}  & \it{358} & \it{359} & \it{344}\\
  \bottomrule
  \end{tabular}
  \label{tab:attacks-distribution}
\end{table}

%------------------- NOT DONE

\begin{figure}[h]
\centering
\includegraphics[width=0.8\linewidth]{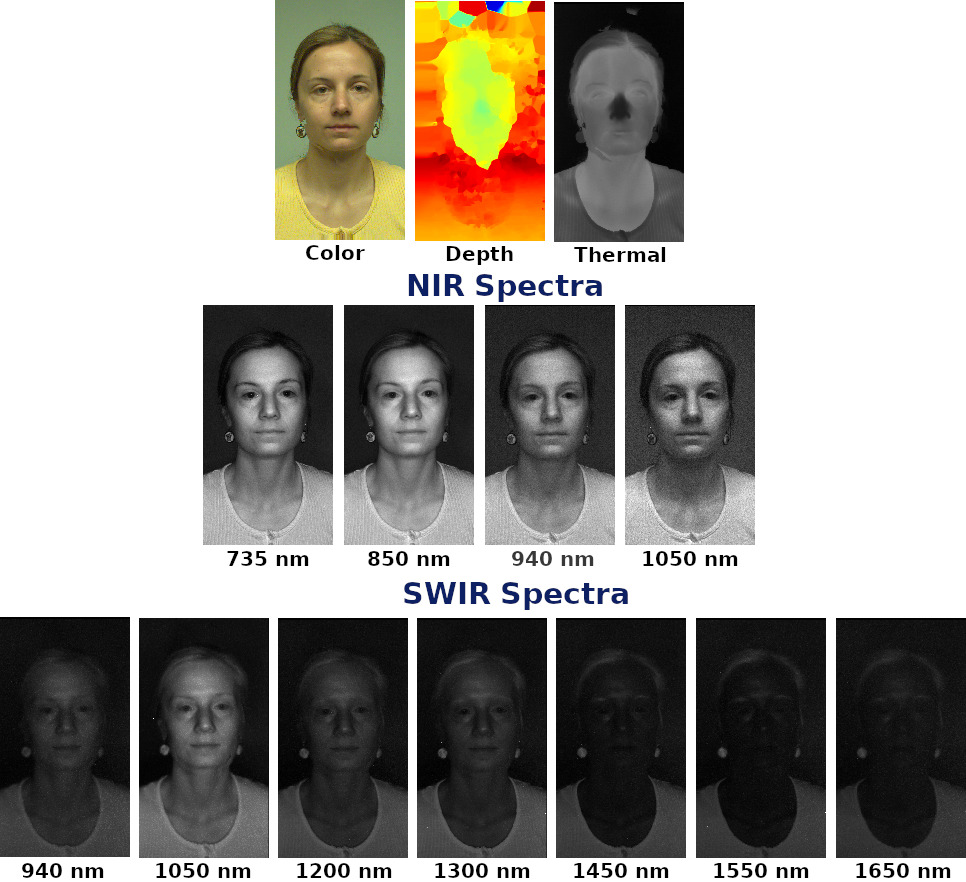} % 0.5 in full
\caption{Different channels from the face images, first row shows the color, depth and thermal channels; second row shows the channel-wise normalized images from four NIR wavelengths, and the third row shows the channel-wise normalized images from different SWIR wavelengths.}
\label{fig:channels}
\end{figure}

Examples of different channels present in the database are shown in Fig. \ref{fig:channels}. Please refer to the reference publication for more information.

\section{Preprocessed files}

Since the original database is large and cannot be downloaded directly, we provide preprocessed files to reproduce the paper. Specifically, we provide two sets of preprocessed files for the two CNN models used in the reference publication. The processed files are in folders \textit{MCCNN-128} and \textit{MC-PixBiS-224}.
The difference between the two sets is in the size of image crop and the type of alignment. All the other steps are similar.

\begin{itemize}
    \item \textit{MCCNN-128}: This is the set of preprocessed files to use with the $MCCNN$ models presented in the reference paper. The images are of size ($128 \times 128$) and the face crop is loose made suitable for the LightCNN model.
    \item \textit{MC-PixBiS-224} : This is the set of preprocessed files to use with the $MC-PixBiS$ models presented in the reference paper. The images are of size ($224 \times 224$) and the face crop is tight as shown in images below.
\end{itemize}

The preprocessing for the color channel is done in several steps. First a face detection is performed. Once the face bounding box is obtained, face landmark detection is performed in the detected face bounding box. Then the images are aligned by transforming them such that the eye centers and mouth center are aligned to predefined coordinates. The aligned face images are converted to grayscale, and resized, to the resolution of 128$\times$128 pixels. The preprocessing stage for non-RGB channels is done by reusing the facial landmarks detected in the color channel and then a similar alignment procedure is performed. The images are then normalized to convert the range of the non-RGB images to 8-bit format.

Instead of directly using images at different SWIR wavelengths, a normalized difference between these images has been considered. This normalization is independent of the absolute brightness and
exhibits differences between skin and non-skin pixels .
Consider two SWIR images of the same individual, $I_{s_1}$ and $I_{s_2}$, recorded at (almost) the same time\footnote{There is a lag of 11ms between frames recorded at different wavelength,
resulting in a total lag of 77ms within the considered SWIR range.} but at different wavelengths, the normalized difference is given by:

\begin{equation}
\label{eq:swir-diff}
  d(I_{s_1}, I_{s_2}) = \frac{I_{s_1} - I_{s_2}}{I_{s_1} + I_{s_2} + \epsilon}
\end{equation}

 $\epsilon$ was set to $1e^{-4}$. Since our recording setup allows to capture SWIR data at no less than $n=7$ different wavelength in each recordings, the number of possible
SWIR image differences is hence given by:

\begin{equation}
\frac{n!}{(n-2)!} = \frac{7!}{5!} = 6 \cdot 7 = 42
\end{equation}

The preprocessed files provided contains a total of 43 channels, i.e., one grayscale image and the 42 combinations of SWIR wavelengths. The details can be found from the paper package.

\begin{figure}[h]
\centering
\includegraphics[width=0.95\linewidth]{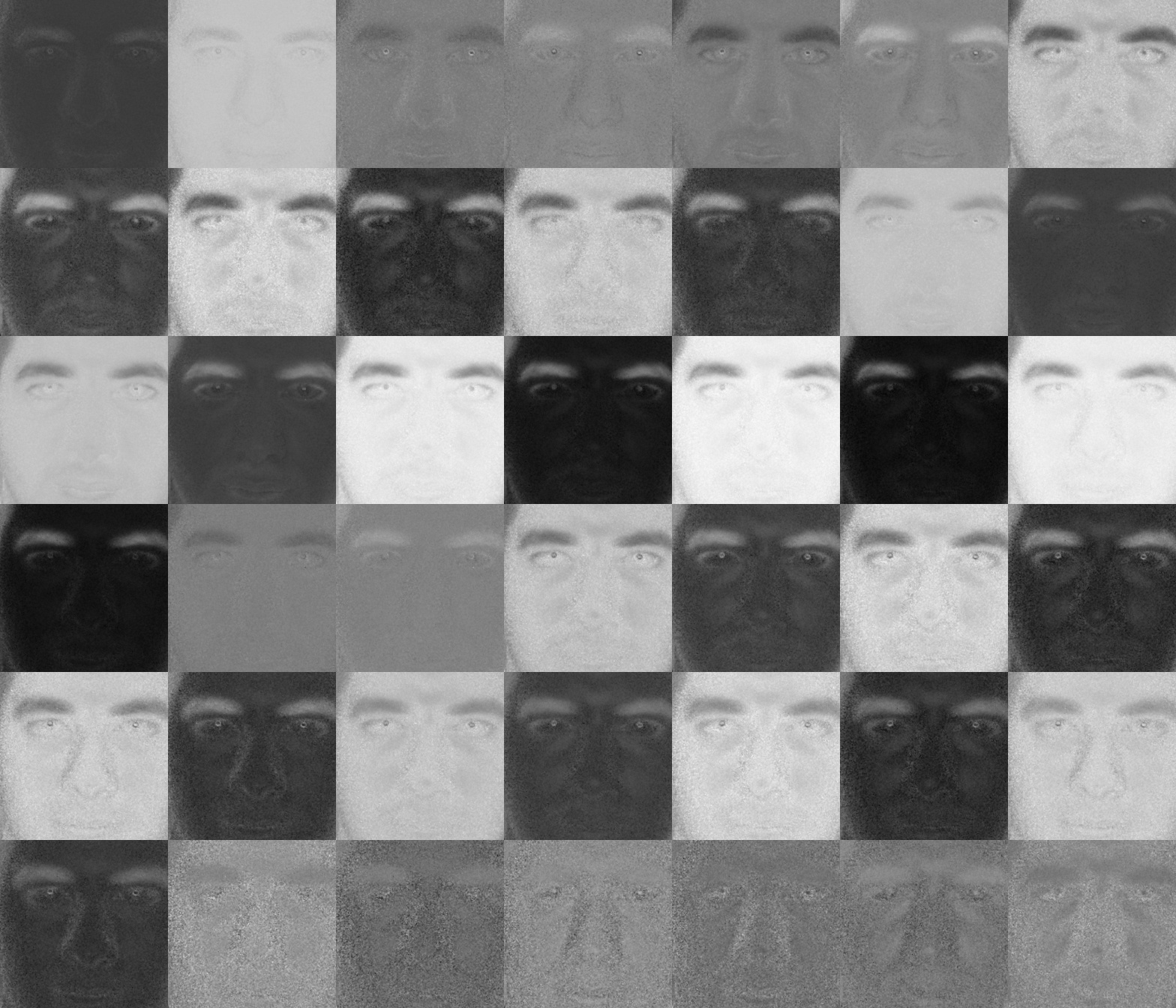} % 0.5 in full
\caption{All 42 SWIR difference images at resolution $224 \times 224$, for the MC-PixBiS Model}
\label{fig:MC-PIXBIS-224-ALL}
\end{figure}

\begin{figure}[h]
\centering
\includegraphics[width=0.25\linewidth]{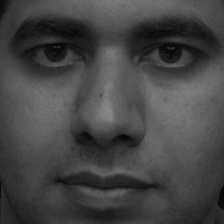} % 0.5 in full
\caption{The grayscale image at resolution $224 \times 224$, for the MC-PixBiS Model}
\label{fig:MC-PIXBIS-224-GRAY}
\end{figure}

\begin{figure}[h]
\centering
\includegraphics[width=0.95\linewidth]{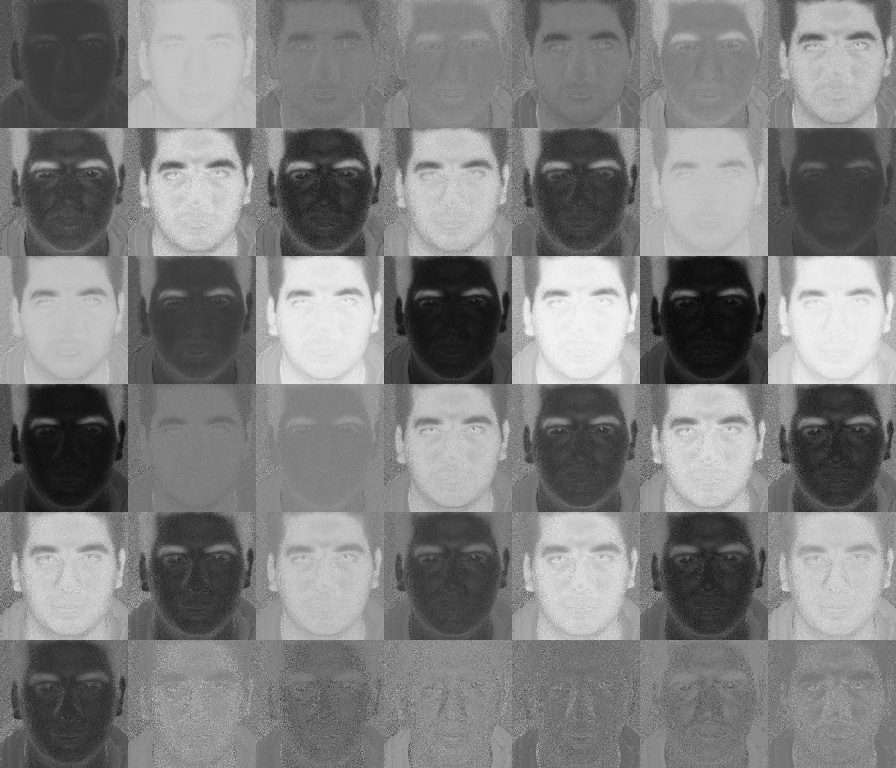} % 0.5 in fullb
\caption{All 42 SWIR difference images at resolution $128 \times 128$, for the MC-CNN Model}
\label{fig:MC-CNN-128-ALL}
\end{figure}

\begin{figure}[h]
\centering
\includegraphics[width=0.25\linewidth]{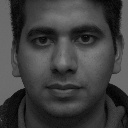} % 0.5 in full
\caption{The grayscale image at resolution $128 \times 128$, for the MC-CNN Model}
\label{fig:MC-CNN-128-GRAY}
\end{figure}

The preprocessed method is described in details in the reference publication and the implementation is available publicly \footnote{\url{https://gitlab.idiap.ch/bob/bob.paper.pad_mccnns_swirdiff}}. The distribution of samples in each set and the details of the protocols can be found from the reference publication and the database package \footnote{\url{https://gitlab.idiap.ch/bob/bob.db.hqwmca}}.

\clearpage

%-------------------------------------------------------------------------
\section*{Acknowledgment}
%-------------------------------------------------------------------------

Part of this research is based upon work supported by the Office of the
Director of National Intelligence (ODNI), Intelligence Advanced Research
Projects Activity (IARPA), via IARPA R\&D Contract No. 2017-17020200005.
The views and conclusions contained herein are those of the authors and
should not be interpreted as necessarily representing the official
policies or endorsements, either expressed or implied, of the ODNI,
IARPA, or the U.S. Government. The U.S. Government is authorized to
reproduce and distribute reprints for Governmental purposes
notwithstanding any copyright annotation thereon.

{\small
\bibliographystyle{IEEEtran}
\bibliography{egbib}
}

\end{document}